# Lithium Metal Battery Quality Control via Transformer-CNN Segmentation

Jerome Quenum[1, 2], Iryna Zenyuk[3], and Daniela Ushizima[1, 2]

[1]University of California, Berkeley
[2]Lawrence Berkeley National Laboratory
[3]University of California, Irvine

jquenum@berkeley.edu,  izenyuk@uci.edu,  dushizima@lbl.gov

## Abstract

Lithium metal battery (LMB) has the potential to be the next-generation battery system because of its high theoretical energy density. However, defects known as dendrites are formed by heterogeneous lithium (Li) plating, which hinders the development and utilization of LMBs. Non-destructive techniques to observe the dendrite morphology often use X-ray computed tomography (XCT) to provide cross-sectional views. To retrieve three-dimensional structures inside a battery, image segmentation becomes essential to quantitatively analyze XCT images. This work proposes a new semantic segmentation approach using a transformer-based neural network called TransforCNN that is capable of segmenting out dendrites from XCT data. In addition, we compare the performance of the proposed TransforCNN with three other algorithms, such as U-Net, Y-Net, and E-Net, consisting of an Ensemble Network model for XCT analysis. Our results show the advantages of using TransforCNN when evaluating over-segmentation metrics, such as mean Intersection over Union (mIoU) and mean Dice Similarity Coefficient (mDSC) as well as through several qualitatively comparative visualizations.

## 1 Introduction

Lithium metal batteries (LMBs) offer high specific energy density, as Li is a light element. Furthermore, the liquid electrolyte is not needed in LMBs and polymer or ceramic electrolytes can be used, which are inherently more safe compared to organic electrolytes. However, currently, no commercially-available LMB systems exist because of Li dendrite formation, resulting from inhomogeneous Li-metal plating. This section describes current methods for imaging batteries followed by a review of the main algorithms for image analysis, semantic segmentation, and battery characterization.

### 1.1 Assessing Battery Quality with Imaging

Synchrotron-based hard X-ray computed tomography (XCT) has spatial resolution suitable to resolve dendrite structures that have microscale dimensions [1, 2]. Lithium has a low atomic number, therefore a low X-ray attenuation. For example, Li dendrites will appear to be void spaces within dense polymer electrolyte materials. XCT imaging of LMB seldom recovers chemical information and relies on differences in material thicknesses and atomic numbers to differentiate Li from solid polymer electrolytes (SPE). To worsen the detection of LMB dendrites, they are porous formations that lead to large intensity variations within their volume. When the LMB is subjected to multiple charge-discharge cycles, a phenomenon known as pitting corrosion develops, and an electrode can present a combination of pits and dendrites, with both presenting similar X-ray attenuation in XCT data. Thus, it is challenging to differentiate dendrites from pits but also to quantify them properly.

Previous LMB *operando* studies using XCT analyzed Li metal plating and how the interphase evolves in symmetric Li-Li cells with polymer electrolytes and in the batteries with Li-metal anode [3, 4, 2, 5, 6]. These studies focused on the battery design and functioning, however, they lack methods for revealing the structure of the dendrites. Most of these previous studies applied traditional thresholding algorithms to conduct segmentation that is unfortunately not reproducible when applied to new samples and rarely applicable to the differentiation between Li metal, pits, and other materials. Alternatively, manual segmentation could be used for a selected cross-section, but it is often unfeasible for full-stack high-resolution imaging surveillance because it is a highly time and labor-intensive process [7].

## 1.2 Deep Learning for Semantic Segmentation

Semantic segmentation is the computer vision task of splitting an image into different categories [8] using a data-driven model to assign a pixel-wise classification given the input image. Different models have been proposed, among them, the Fully Convolutional Networks (FCN) [9], which proposed a paradigm shift in 2015 that run fully connected layers and includes a way to allow for each pixel to be classified from feature maps coming from convolutional layers. This builds on a series of local convolutions of preceding layers that aim to obtain a representation of multi-scale feature maps used for the classification tasks. Around the same time, Ronneberger et al. [10] introduced their convolutional neural network (CNN), a new CNN-based encoder-decoder model known as U-Net, and showed that combining higher and lower features symmetrically is beneficial in obtaining better performance. Soon after, SegNet [11] and Deeplab[12] were proposed, confirming that the encoder-decoder architecture is well suited for such a task.

A lot of these works [12, 13, 14, 15] also leverage the *atrous convolution* to show that it could help capture contextual information. In particular, Y-Net [15] used three modules to improve segmentation accuracy. In addition to the *Regular Convolution Module*, and the *Pyramid Pooling Module* which allows the model to learn global information about potential locations of the target at different scales, the *Dilated Convolution Module* took advantage of the fact that the target is often shared out in the samples, which supports learning sparse features in their structure. PSPNet[14] on the other hand, adds ResNet[16] as a backbone while multi-scale feature maps are aggregated in its encoder.

Though these architectures work well, the computer vision community has increasingly seen their design shift from pure CNN-based design with [9, 10, 11, 12, 14, 15] to transformer-based designs which started with **ViT** [17, 18, 19, 20, 21]. Later on, hybrid models started exploiting the best of both worlds by either using a transformer as encoder and CNN as decoder [22, 23], or CNNs as encoder and a transformer as decoder, or even using CNNs and encoder-decoder while transformers are used in the middle or in between to process the feature maps. One such hybrid model is HRNet-OCR [24] which has a CNN as a backbone and combines it with cross-attention layers between features of different scales to account for multiple contexts and scales in the data.

In all these schemes, the common denominator remains the **attention mechanism** which has proven in the past few years to exceed the performance of models disregarding it. That is because it allows the features not to be subject to the inductive biases and translation invariance that occurs in CNNs. Instead, it allows the model to learn long-term dependencies between pixel locations [17]. In other words, it allows for a better representation by leveraging **contextual information** either between pixels, patches, or channels.

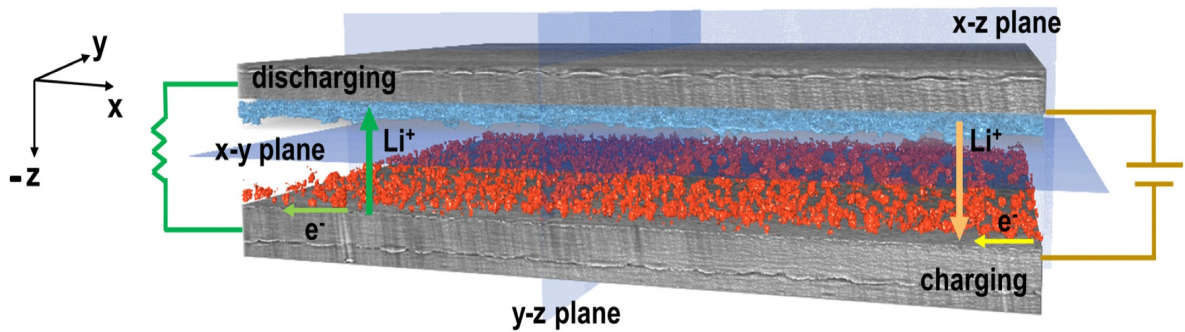

Figure 1: Diagram illustrating the Li–polymer–Li symmetric cell design, imaged using X-ray CT, with highlighted dendrite formations (blue) and the redeposited Li (red).

## 1.3 Problem and Motivation

Li dendrite formations initiate during battery cycling as illustrated in Figure.1, with dendrites nucleating on the interface between the electrolyte and the electrodes. Dendrite growth depends on the current density of plating, electrolyte transference number, electrolyte mechanical properties, and impurities present in Li-metal material and at the interfaces. Earlier works have shown that increasing the shear modulus of the SPEs can help suppress Li dendrite growth but cannot fully eliminate it [25, 26, 27, 28]. Further studies have shown that Li-metal surface impurities ($Li_3N$, $Li_2CO_3$ and $Li_2O$) can result in inhomogeneous current density and promote nucleation of Li dendrites.

Accurate segmentation for measuring dendrite volume has guided research and quality control of battery designs as well as tests of materials used for its components. Deep learning methods can provide exceptional segmentation results [29, 30, 31] when using high-resolution XCT data, particularly when large collections of annotated data are available. For example, Zhang et al [32] used a CNN known as D-LinkNet to inspect the effect of distortion on the segmentation accuracy of Li-ion batteries. Additional works by the same team [33] used a U-Net for multiphase segmentation of battery electrodes from nano-CT images. Despite being focused on battery segmentation, those studies lack information on dendrite segmentation.

Previous studies [34, 35] on inspecting dendrites in batteries discussed problems regarding the mechanisms and types of nucleation, e.g., lateral growth or Li filaments. Data acquisition modes range from electron microscopy [34, 36, 37] to XCT [35, 37] with valuable morphological characterization and designs for suppression of dendrite growth, but dendrite detection was addressed mostly qualitatively through dendrite projections and/or visualizations. For example, the dendrite volume calculation in [37] was based on median filter and Otsu thresholding, a method that seldom works for more than a few slices from an XCT stack, unless considering strenuous manual post-processing [38].

## 1.4 Research Contributions

The proposed work describes the design and implementation of dendrite and Li deposit segmentation from 3D XCT images. In performing the segmentation of a 3D volume automatically, we could either design a 3D model that performs directly on an input volume or we could leverage 2D models by subdividing the volume into slices. This paper introduces a 2D model due to its ability to be trained faster and with a limited number of samples, hence making it more versatile. In particular, we propose an architecture that benefits from both the contextual information learned from transformers and the global information captured by CNNs to predict dendrites and Li deposits from a lithium metal battery that underwent cycling, and was imaged using high-resolution XCT data. This article compares four different deep learning architectures, including U-Net, Y-Net, TransforCNN, and E-Net, on their ability to segment dendrites inside the cycled symmetric Li-Li battery with polymer electrolytes.

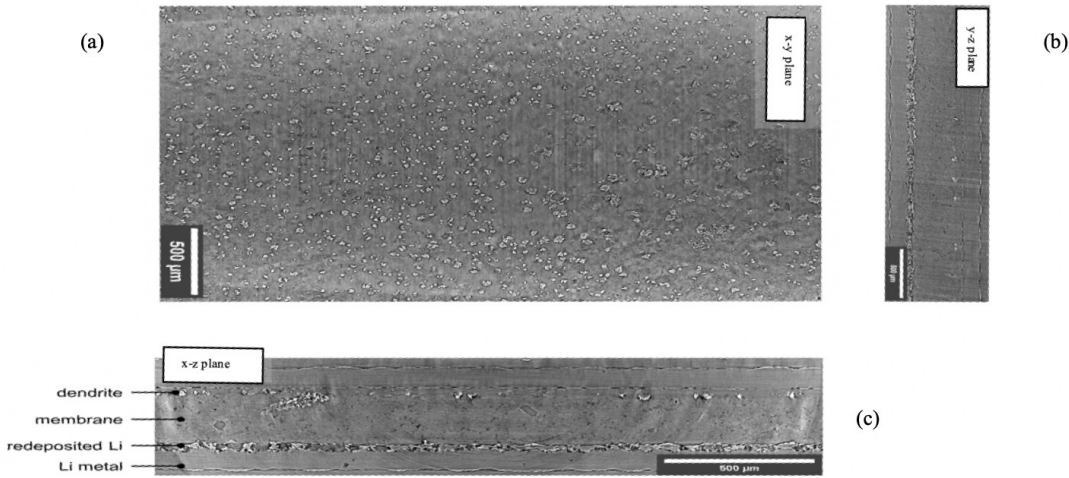

Figure 2: Cross-Sectional Images for the Li–polymer–Li symmetric cell; (A) Cross section of the x-y plane where the training was done on this plane; (B) Cross-sections of the x-z plane and detailing of the cell components; (C) Cross-sections of the y-z plane.

## 2 Materials Description

Li metal holds a high theoretical capacity (3860 mAh/g) and a large negative thermodynamic potential (-3.06 V vs. SHE)[39]. Thus, it is considered a promising candidate for the next-generation battery anode. Li metal is highly active and can introduce a series of side reactions in a battery system with liquid electrolytes. This can also cause the dendrite to form, which would eventually lead to short circuits and bring safety issues to the battery. Using solid electrolytes, instead of liquid electrolytes, a more stable interface can be designed between the electrolyte and the Li metal electrodes, thus alleviating the dendrite formation issue. Recent developments in electrolyte engineering can be found in [40, 41, 33, 37].

Currently, several solid-state materials are used as electrolytes and separators, such as polymers and single-ion conducting inorganic solid electrolytes (glass or ceramic). Polymer materials are promising as they are mechanically flexible because polymer materials can be produced in a roll-to-roll scalable process and be designed very thinly. However, for SPEs to have broad deployment, strategies for dendrite suppression must be developed, such as coatings and soft interlayers including polymers as well as ionic liquids [42, 43, 44].

Design of interfaces in Li-metal batteries or All Solid-State Batteries (ASSBs) is challenging, as it involves control of Li-ion plating onto Li metal and this plating process needs to be uniform. When not done properly, interactions at the interfaces of electrodes arise due to transport processes associated with ions [45, 41], which can form ionic aggregates. Dendrite growth can be triggered by the formation of localized regions of high lithium ion concentration, which can occur due to the clustering of lithium ions into ionic aggregates. In general, dendrites are formed during battery charge and discharge cycles. This happens especially when a battery is charged at high current densities, due to the heterogeneous Li metal plating even when considering solid electrolytes. Tracking the structure and evolution of dendrites is important to develop a strategy to prevent their growth. Dendrites are tree-like and porous structures, usually with a size in nano to micro-scale. Given the morphological structure of dendrites and their size relative to an input stack, performing XCT segmentation is a suitable method of analysis as it allows for a pixel-wise classification which helps quantify the volume of dendrites and use it as a proxy of battery quality.

## 2.1 Electrochemical Testing

Li/Li symmetric cells are assembled using two Li metal electrodes and are considered a tool for testing and observing the Li metal anode without being affected by cathode materials.

Free-standing Li-metal foils with a thickness of 100 µm from FMC were used. Polymer electrolyte membrane was sourced from an industrial partner with a thickness of 140 µm as a research sample. The cell was assembled as Figure. 3 shows. A red shim with a thickness of 50 µm was used to create a circle with a 0.8 cm diameter. Two circular polymer electrolytes with a diameter of 0.80 cm were punched out and placed on each side of the red shim. Two Li-metal foils were then placed on the outside of the membranes as electrodes. The electrodes were connected to the metal tabs. The cell was sealed with a vacuum sealer. A current density of 1.5 mA/cm$^2$ was periodically applied to the cell for 10 minutes, and the battery rested for 20 minutes. The cell was cycled at 3.0 mAh/cm$^2$ for one full cycle (120 minutes for charging and 120 minutes for discharging), after which the cell XCT scan was acquired.

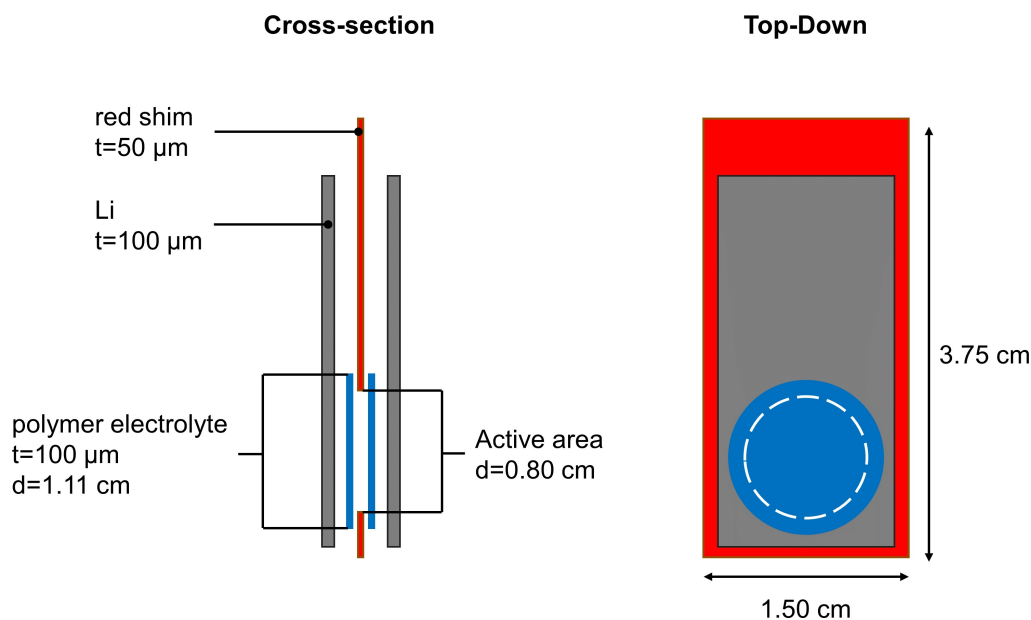

Figure 3: Schematic illustration of the pouch cell.

## 2.2 Synchrotron X-ray CT Imaging

The XCT scan was acquired at Beamline 2-BM at Advanced Photon Source (APS) at Argonne National Laboratory (ANL), which used a 20 µm LuAG scintillator with 5× lenses, and an sCMOS PCO edge camera. A 27.5 keV energy was selected using a multilayer monochromator, with 100 ms exposure time per back-projection and over 180 degrees of rotation enabling the collection of 1,500 projections. The yielded image presents a resolution of 1.33 µm/pixel and a field-of-view of 3.3 mm. Three FOVs were recorded and were stitched together to form a vertical height of >3 mm during the post-processing. Tomographic reconstructions considered TomoPy with Gridrec algorithm[46, 47, 48].

## 2.3 Raw data Pre-Processing

The resulting raw TomoPy reconstruction was a large volume of size (3977, 2575, 2582) and was not properly aligned as expected due to the various motion involved in collecting the data as shown in Figure. 4. As this raw data contains a lot of noise and irrelevant parts, we proceed to develop an algorithm that will allow us to cleanly crop out those regions. In the process, we inverted the grayscale volume and obtained a maximum projection image on the stacks to facilitate our ability to locate corners.

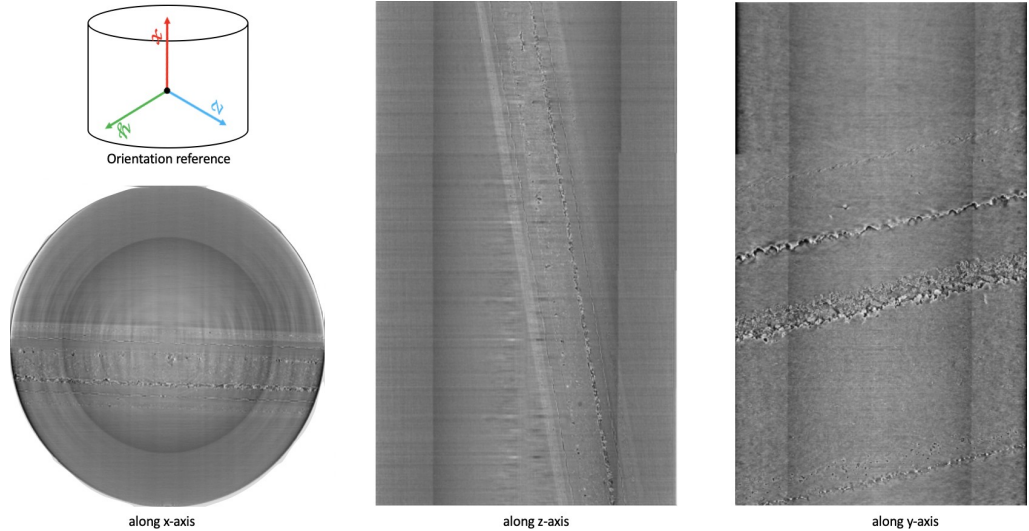

Figure 4: Sample Raw data obtained after TomoPy reconstruction of a CT scan

To align the data, we first used a series of perspective transformations and homography to rectify the region of interest along each plane. Though this process could be automated using feature detectors and feature matching techniques such as MOPS [49] and SIFT [50], we manually selected corners for optimal precision. We then rectified and cropped the raw data to obtain the region of interest to a volume of size (3849, 340, 2071) shown in Figure. 5.

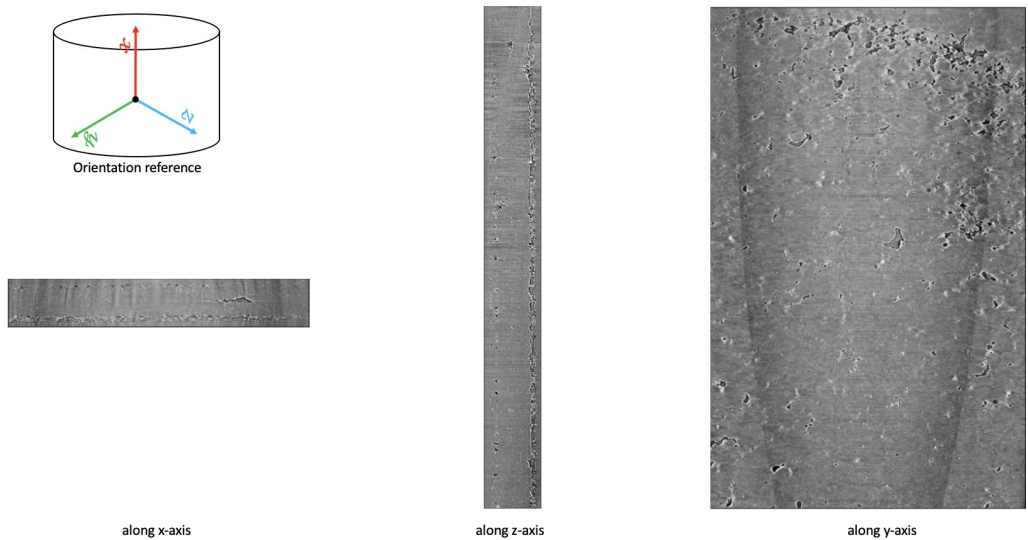

Figure 5: Sample Region of Interest (RoI) data obtained after pre-processing TomoPy reconstruction of a CT scan

## 3  Computational Methods

For a given volume stack, we apply 2D models due to their versatility. In doing so, we subdivide each training hand-labeled slice into 128 × 128 patches which were then separated into training, validation, and testing datasets. Based on their known performances over the years, we investigate CNN-based architectures such as U-Net [10] and Y-Net [15]. In addition, we also compared their performance with TransforCNN, our **proposed Transformer encoder-based network**, and E-Net, an Ensemble Network over U-Net, Y-Net, and TransforCNN. We train the networks in a weakly-supervised fashion where a small subset of labeled data was used in conjunction with a much larger unlabeled sample size. Figure. 11 shows a sample output of the model considered in this work.

## 3.1 U-Net

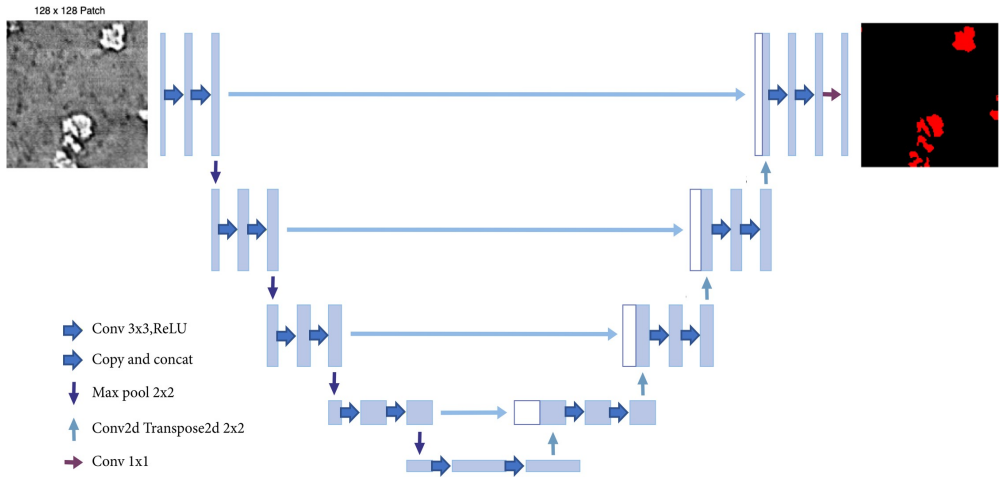

Figure 6: U-Net Architecture.

U-Net is a CNN architecture that has first been introduced in 2015 by Renneberger et al. for the semantic segmentation of biomedical images. We refer interested readers to [10] for details about the architecture of the network. In this work, the model was adapted to take images of size $128 \times 128$ as input. For the encoder, we started with 16 - channel $3 \times 3$ kernels and doubles the number at each layer, followed by a ReLU activation and max pooling until a 256 - channel $8 \times 8$ resolution feature map is obtained. We reversed the operation with transposed convolutions operation on the decoder side and concatenate with corresponding size encoder feature maps until the desired output shape is obtained.

## 3.2 Y-Net

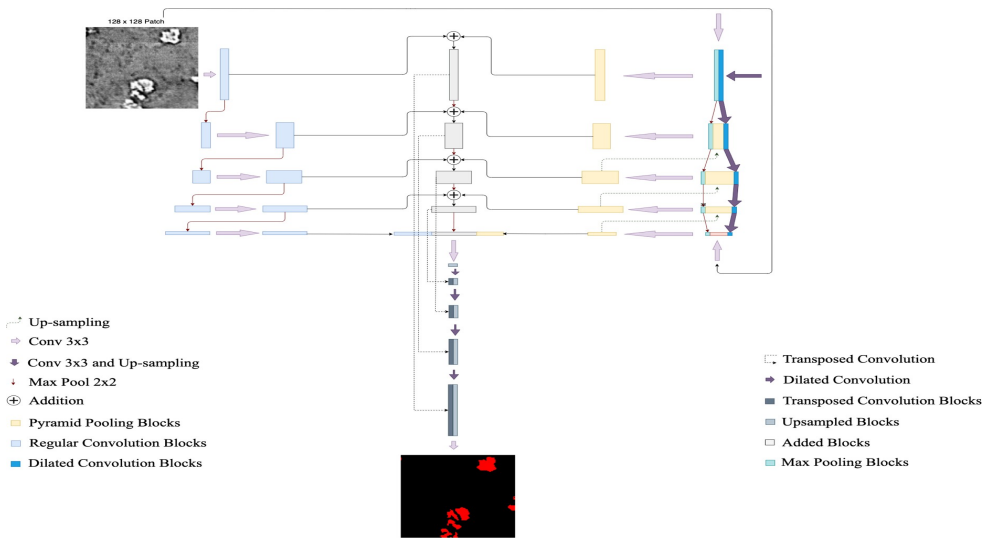

Figure 7: Y-Net Architecture.

Y-Net is a CNN architecture originally introduced by Quenum et al [15] to segment barcodes from Ultra High-Resolution images. By leveraging Y-Net's architecture, we have modified and adapted the *Regular Convolution Module* to take in $128 \times 128$ images from training slices. As it consists of convolutional and pooling layers, we started with 24 - channel $3 \times 3$ kernels and doubled the number at each layer. We alternated between convolution and max-pooling until we reached a feature map size of $8 \times 8$ pixels. The *Dilated Convolution Module* here took advantage of the fact that dendrites are often shared out in the samples to learn sparse features in their structure. It also took $128 \times 128$ input patches and we maintained 16 − channel 3 × 3 kernels throughout the module while the dimensions of the layers were gradually reduced using a stride of 2 until a feature map of 8 × 8 pixels is obtained. Finally, the *Pyramid Pooling Module*, which allows the model to learn global information about potential locations of the dendrites at different scales, had its layers concatenated with the layers on the dilated convolution module to preserve the features extracted from both modules.

### 3.3 TransforCNN

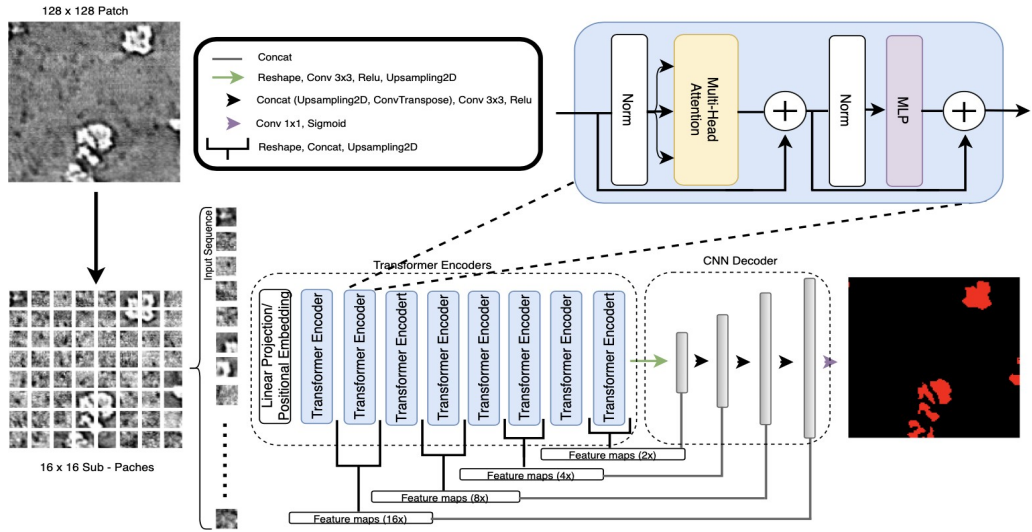

Figure 8: T-Net Architecture.

TransforCNN is a hybrid Transformer-CNN segmentation model that leverages the encoder model of Vision Transformers ViT [17] and the decoder architecture of CNNs. More specifically, its encoder model was first introduced in Natural Language Processing (NLP) by Vaswani et al [51] and its multi-headed self-attention was later shown (by ViT) to help remove the common inductive biases observed in CNN-only models by relating all input sequence with each other. As depicted in Figure. 8, the proposed architecture is hybrid because it combines the Transformer Encoders Block with the CNN Decoder Block to deliver semantic segmentation.

The ***Transformer Encoders Block*** takes inputs that are 16×16 sub-patches sequences from the 128×128 patches that were obtained from training slices. These patches are flattened and each is embedded into a 64-dimensional feature vector via a linear projection and is added to its corresponding Fourier Features (FF) positional encoding. We used 8 transformer encoder units and the outputs of every 2 transformer encoders were reshaped into a 2-dimensional feature map representation, concatenated, up-sampled recombined with the layers from the CNN Decoder Block of corresponding dimension.

The ***CNN Decoder Block*** takes in the output of the last transformer encoder unit, reshapes it into a 2-dimension representation on which a set of 3 × 3 kernels convolutions and max-pooling is applied to obtain a feature map of 8 × 8 pixels. The resulting feature maps are then concatenated with corresponding size feature maps coming from the Transformer Encoder Block and up-sampled continuously until the final output is obtained. This last step allows for the enhancement of the features in the CNN Decoder Block as we are progressively reconstructing the output dimension of 128 × 128.

### 3.4 E-Net

We have combined the results of different architectures, namely U-Net, Y-Net, and TransforCNN, to create an ensemble prediction scheme called E-Net. It was found that our best mean Intersection over Union (mIoU) is obtained when combining 20% of U-Net with 80% of TransforCNN while the best mean Dice Similarity Coefficient (mDSC) is obtained using only a TransforCNN.

## 4 Experimental Results

In training the models (U-Net, Y-Net, and TransforCNN), we used one NVIDIA Tesla V100 GPU for each experiment. We obtained a total of 4433 samples of resolution 128 × 128 with their corresponding hand-labeled ground truth that the models were trained on. We used 80% of the examples for the training set, 10% for the validation set, and 10% for the testing set. We used data augmentation schemes in training all models which consist of random rotations in all directions, random flips (vertically and horizontally), random cropping (2%), random shifts, random zoom (range in [0.8, 1]), and a small range of random brightness and contrast variation (+/_ 5%). We trained the U-Net for 450 epochs while the Y-Net and TransforCNN models were trained for 130 and 300 epochs respectively.

As shown in Figure. 4, the Y-Net and TransforCNN converge faster than U-Net with the initial loss of the TransforCNN model being significantly lower than that of the U-Net and Y-Net models. We have experimented with various loss functions such as Tversky loss [52] described in Eq. 1, the focal Tversky loss [53] described in Eq. 2, the balanced cross-entropy loss described in Eq. 4, and the binary cross-entropy loss (Eq. 3) out of which the latter yields the best results. One interesting observation is that though the validation curve on U-Net exhibits characteristics of a better generalization, the quantitative results show otherwise.

For evaluation, we have used the dice similarity coefficient described in Eq. 5 and the Jaccard Index also known as Intersection over Union Eq. 6. In all the equations, $y$ and $\hat{y}$ are respectively the ground truth and prediction on patch $i$, and TP, FP, and FN represent the number of true positives, false positives, and false negatives respectively.

$$L\_Tversky(y, \hat{y}) = \frac{y\hat{y}}{y\hat{y} + \beta(1-y)\hat{y} + (1-\beta)y(1-\hat{y})} \quad (1)$$

where $\alpha, \beta > 0, \alpha + \beta = 1$.

$$L\_Focal\ Tversky(y, \hat{y}) = \left(1 - L\_Tversky(y, \hat{y})\right)^\gamma \quad (2)$$

where $\alpha, \beta > 0, \alpha + \beta = 1, \gamma = 4/3$.

$$L\_Cross\ Entropy(y, \hat{y}) = -y \log(\hat{y}) - (1-y)\log(1-\hat{y}) \quad (3)$$

$$L\_balanced\ Cross\ Entropy(y, \hat{y}) = -\beta y \log(\hat{y}) - (1-\beta)(1-y)\log(1-\hat{y}) \quad (4)$$

where $\beta \in [0, 1]$.

$$DSC(y, \hat{y}) = \frac{2TP}{2TP + FP + FN} \quad (5)$$

$$IoU(y, \hat{y}) = \frac{TP}{TP + FP + FN} \quad (6)$$

$$(7)$$

Note that Eq. 6 could also be expressed as DSC / (2-DSC).

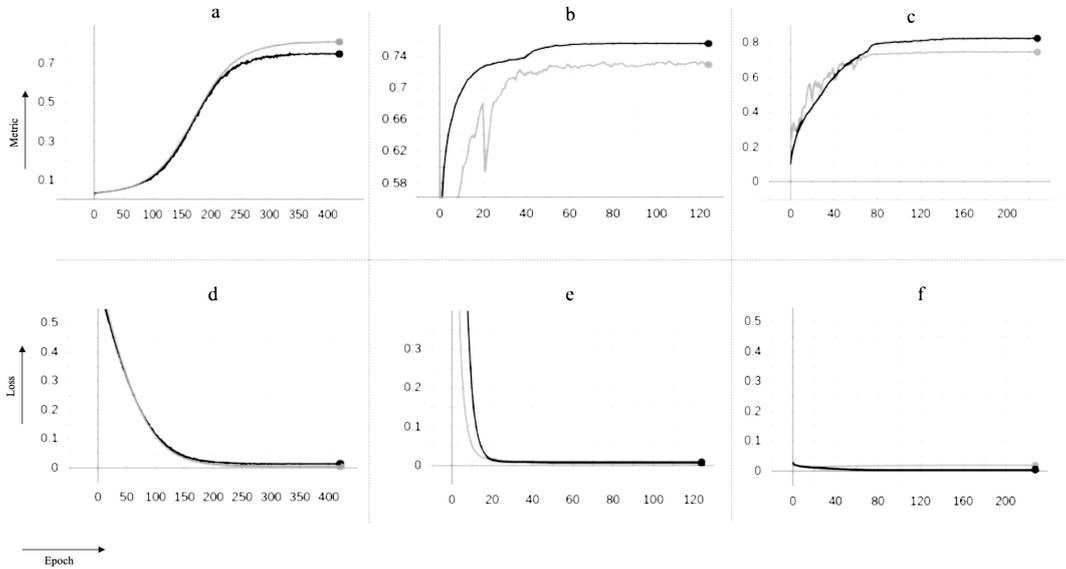

Figure 9: Training curves for U-Net, Y-Net, and TransforCNN on the y-axis vs numbers of epochs on the x-axis; the models were optimized over the binary cross-entropy function as loss and evaluated on the dice similarity coefficient as evaluation metric during training; The gray curves depict behavior on the validation sets while the black curves show behavior on the training sets over increasing numbers of epochs; (a) training and validation dice coefficient for U-Net; (b) training and validation dice coefficient for Y-Net; (c) training and validation Dice coefficient for TransforCNN; (d) training and validation loss for U-Net; (e) training and validation loss for Y-Net; (f) training and validation loss for U-Net; the use of dropout during only the training phase explains why the models tend to perform better on the validation set over time.

|        | mIoU  | mDSC  | latency (ms) | Input Patch Resolution (px) |
|--------|-------|-------|--------------|-----------------------------|
| U-Net  | .8698 | .8998 | **65.36**    | 128 × 128                   |
| Y-Net  | .8481 | .8790 | 103.62       | 128 × 128                   |
| T-Net  | .9511 | **.9647** | 206.75   | 128 × 128                   |
| E-Net  | **.9514** | .9641 | 473.59   | 128 × 128                   |

Table 1: mIoU, mDSC, Inference Time, and Patch Size for Li-Li Symmetric battery Dataset.

|            | Volume (Cubic Px) | Occupied Volume Percentage |
|------------|-------------------|----------------------------|
| U-Net      | 54940997.0        | 2.027                      |
| Y-Net      | 99389447.0        | 3.667                      |
| TransforCNN| 82892014.0        | 3.058                      |
| E-Net      | 80858216.0        | 2.983                      |

Table 2: Volume and Percentage Volume Occupied for U-Net, Y-Net, TransforCNN, and E-Net.

For all of the models, we use the mean Dice Similarity Coefficient (mDSC) and the mean Intersection over Union (mIoU) as metrics shown in Table 1. As seen, our proposed pipeline outperforms U-Net [10], and Y-Net [15] by a *mIoU* of 8.13% and 10.3% and *mDSC* of 6.49% and 8.57% respectively. Also shown in Table 1, is a slight *mIoU* improvement of 0.03% by our Ensemble Network analysis (E-Net) on TransforCNN.

In addition, Table 1 displays that while TransforCNN is successful in segmenting out dendrites, its latency is at least 3.16× slower than U-Net which has the fastest latency of all models evaluated at 65.36 milliseconds (ms). The slowest of all the models is observed to be E-Net which performs 7.24× slower than U-Net.

Qualitatively, Figure. 10 shows the predictions on a given test slice and more specifically, Figure. 11 shows sample predictions at the patch level. As observed in the first and second rows (a;b), the TransforCNN and U-Net predictions are the closest to the Ground Truth. The third, fourth, and eighth rows (c; d; h) show that U-Net and Y-Net tend to generalize better as the unlabeled dendrite regions in the input patches are segmented out by these two models while TransforCNN and E-Net still reflect the Ground Truth images. The fifth, sixth, and seventh rows (e; f; g) show the generalization potential of all models while the predictions of TransforCNN and E-Net overall tend to remain closer to the Ground Truth. In addition, Table 2 summarizes the absolute number of voxels corresponding to the segmented dendrite and re-deposited Li volume as well as their volume fraction.

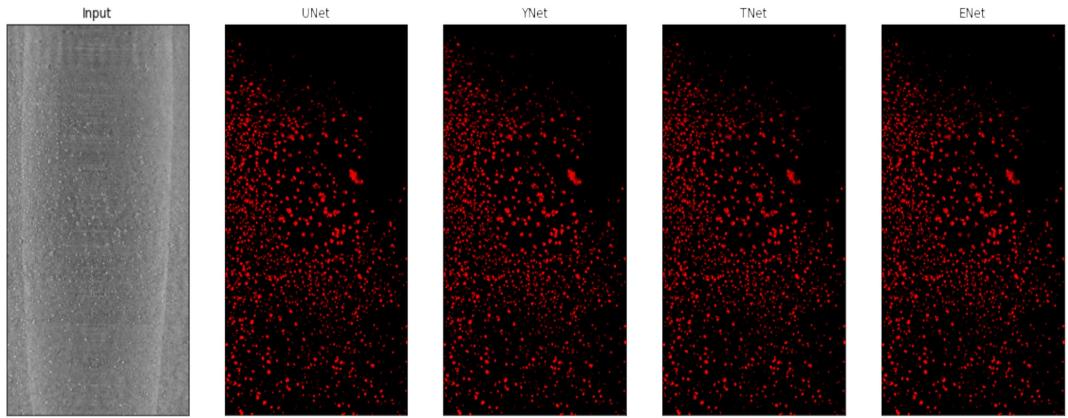

Figure 10: U-Net, Y-Net, T-Net, and E-Net sample outputs on a slice showing a cross-section along the x-y plane

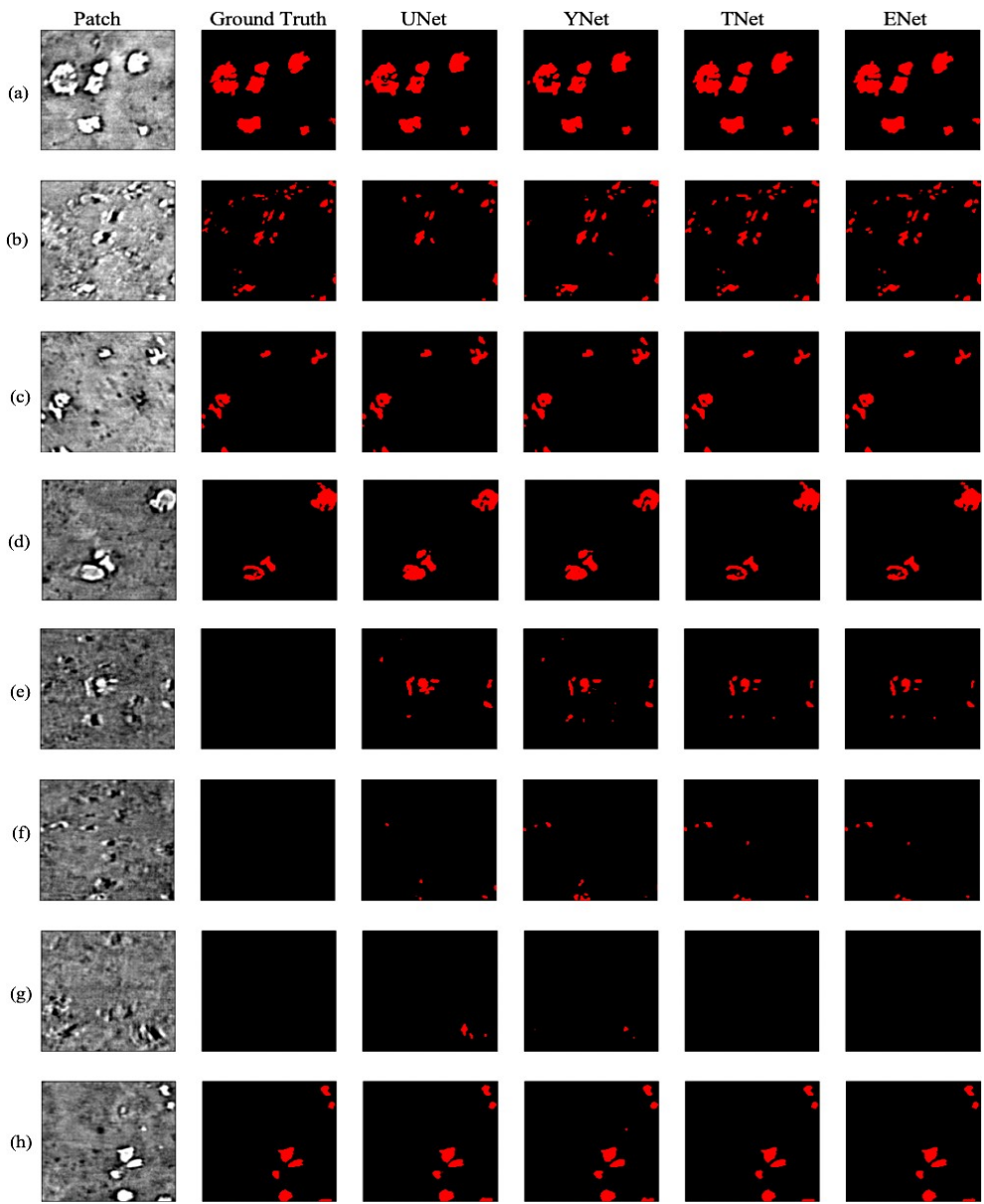

Figure 11: U-Net, Y-Net, T-Net, and E-Net sample outputs on random test patches.

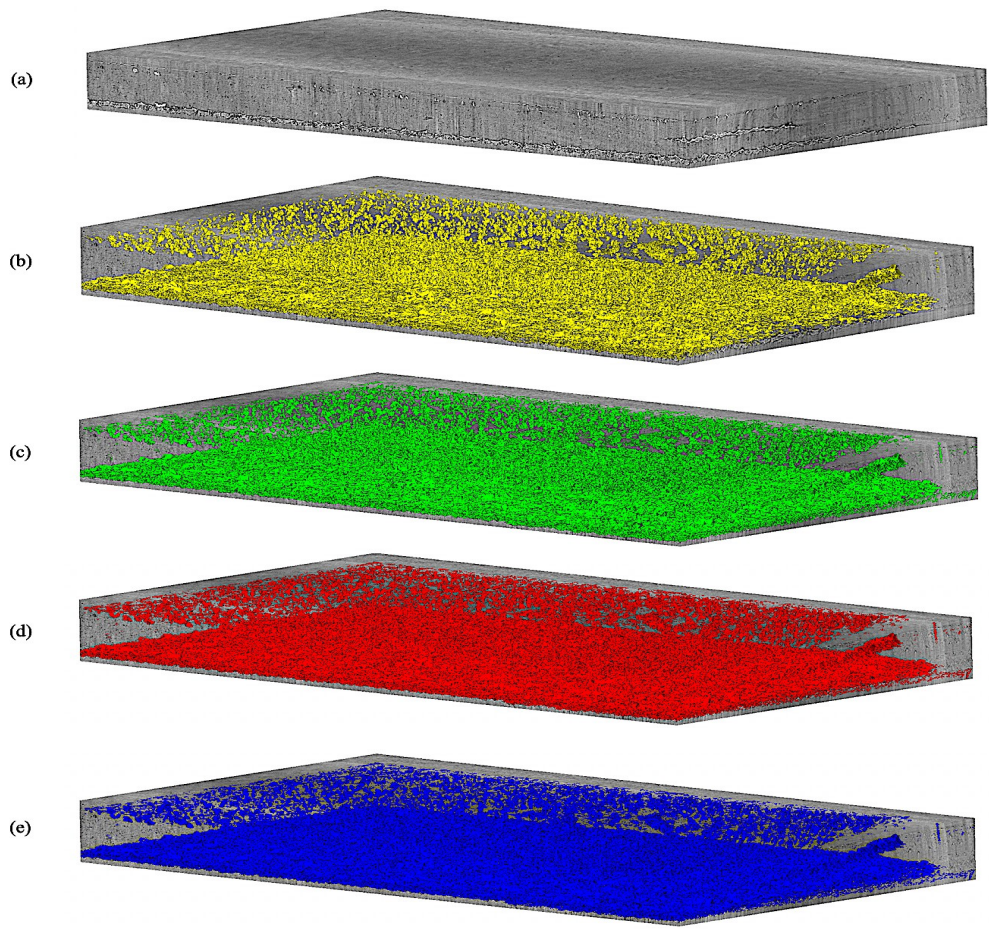

Figure 12: 3D rendering of U-Net, Y-Net, T-Net, and E-Net on test volume; (a) grayscale test input volume; (b), (c), (d), (e) are respectively U-Net, Y-Net, T-Net, and E-Net predictions

# 5  Discussion and Conclusion

LMBs are promising candidates for next-generation batteries because of their high specific energy density. Currently, Li dendrites growth is an issue, as it can lead to loss of Li (dead Li), shorting of cells, and other undesirable degradation phenomena.

More specifically, short-circuiting is a leading failure mechanism in LMBs due to the uncontrolled propagation of lithium protrusions that often present a dendritic morphology. The energy density benefits of using LMBs can only be harvested after scientists are able to detect and regulate the dendrite formation and control dendrite growth.

Uneven lithium ion distribution and dendrite formation can compromise battery performance and safety. For this reason, this paper introduced a semantic segmentation algorithm called TransforCNN that detects both dendrites and re-deposited Li accurately, and compared it with traditional approaches.

Overall, it was observed that TransforCNN and E-Net tend to learn semantics in the Ground Truth images provided during training. In contrast, U-Net and Y-Net tend to simply generalize even to cases where segments in the Ground Truth were wrongly hand-labeled. We speculate that this may lead to U-Net and Y-Net being wrongly penalized during the evaluation process while TransforCNN and E-Net are rewarded since their predictions always look the closest to the Ground Truth.

Experiments have also illustrated that our approach outperforms existing methods though it is slower than the fastest (U-Net) of all considered models. In future work, we aim to extend this method to a multi-class segmentation task, differentiating dendrites from pits, and bubbles while improving the current latency in a weakly supervised fashion.

# 6  Acknowlegement

We thank Pavel Shevchenko and Francesco De Carlo from the Advanced Photon Source, Argonne National Laboratory for their help in obtaining the dataset used in this work. For the imaging work at APS, we acknowledge the Synchrotron as this research used resources of the Advanced Photon Source, a U.S. Department of Energy (DOE) Office of Science User Facility operated for the U.S. DOE Office of Science by Argonne National Laboratory under contract No. DE-AC02-06CH11357. In addition, we thank Ying Huang for generating Figure.3 and for using a multi-class experimental version of TransforCNN trained by Jerome Quenum to render Figure.1. Also, we acknowledge Dula Parkinson from the Advanced Light Source, Lawrence Berkeley National Laboratory for sharing data storage resources deployed by the National Energy Research Scientific Computing (NERSC) facility.